\documentclass{article}
\usepackage[final]{neurips_2026}

\usepackage[utf8]{inputenc}
\usepackage[T1]{fontenc}
\usepackage{hyperref}
\usepackage{url}
\usepackage{booktabs}
\usepackage{amsfonts}
\usepackage{amsmath}
\usepackage{amssymb}
\usepackage{nicefrac}
\usepackage{microtype}
\usepackage{xcolor}
\usepackage{graphicx}
\usepackage{subcaption}
\usepackage{multirow}
\usepackage{enumitem}
\usepackage{bm}
\usepackage{caption}

\newcommand{\ris}{\text{RIS}}

\newcommand{\E}{\mathbb{E}}

\makeatletter
\providecommand{\@trackname}{} 

\AtBeginDocument{%
\let\@oldmaketitle\@maketitle
\renewcommand{\@maketitle}{\@oldmaketitle
  \begin{center}
    \includegraphics[width=0.78\linewidth]{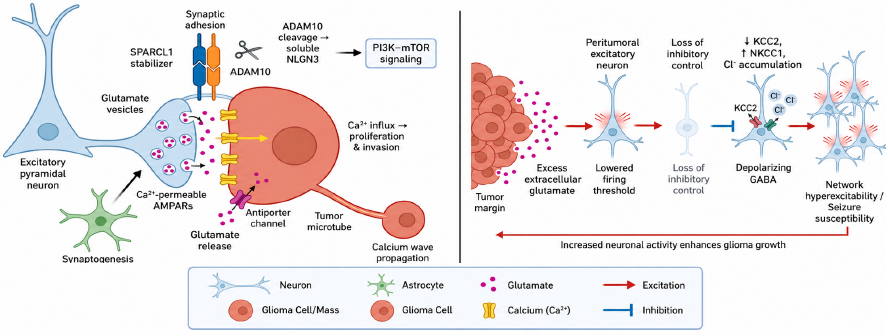}
    \captionof{figure}{\textbf{Neuron--Glioma Synapse formation and the HEx loop.} Glioma cells form functional synapses with neurons via calcium-permeable AMPA receptors. ADAM10-mediated cleavage of NLGN3 activates PI3K-mTOR signaling. Tumor microtubes propagate calcium waves; excess glutamate and altered chloride homeostasis create a feed-forward hyperexcitability(HEx) loop. Unlike the coding machinery the \emph{noncoding regulatory programs} that activate it remain unknown.}
    \label{fig:synapse_schematic}
  \end{center}
  \vspace{2mm}
}%
}
\makeatother

\title{The Dark Regulome: Disentangling Predictability from Regulation in Genomic Foundation Models}

\author{%
\small
  Chahat Baranwal* \\
  IIT Jodhpur \\
  \texttt{b22bb014@iitj.ac.in} \\
  \And
  Aaditya Baranwal* \\
  University of Central Florida \\
  \texttt{aaditya.baranwal@ucf.edu} \\
  \And
  Lakshya Nitin Tandon \\
  Northeastern University \\
  \texttt{tandon.l@northeastern.edu}
}

\begin{document}

\maketitle

\begin{abstract}
High-grade gliomas integrate into neural circuits through functional synapses with neurons, raising the question of which noncoding elements shape synaptogenic gene expression in tumor cells. The regulatory program written across the dark genome, what we call the \emph{dark regulome}, is the natural substrate to probe, and sequence foundation models offer a zero-shot route through in-silico mutagenesis (ISM); yet likelihood-based scoring is tautologically coupled to local sequence predictability, leaving the regulatory interpretation underdetermined. Across three architecturally distinct foundation models (Caduceus-Ph, HyenaDNA, Enformer) and 30{,}448 dark genome elements at 92 glioma-relevant loci, we introduce a residualization-and-permutation diagnostic that separates predictability-driven from regulation-driven RIS variance. A sharp 10\,kb proximal-regulatory horizon survives every control we apply, but the LM-derived element-class hierarchy does not: a six-feature linear baseline matches Caduceus top-decile membership at AUC $= 0.985$. Cross-architecture decomposition cleanly separates a sequence-predictability layer (the two language models co-rank long well-predicted transposable elements) from a regulatory-output layer (Enformer alone retains residual cCRE-discriminative signal), with literally zero overlap between the two top-100 lists. Conservation, brain cis-eQTL, and STRING-PPI cross-checks then anchor what biology survives: top-100 elements across all three models are $3.3\times$ enriched per model for matching brain eQTLs ($p_\mathrm{emp} < 5\times 10^{-3}$), while a tempting transposable-element regulatory layer and a striking NRXN1+NLGN1 protein-pair convergence both fail proper permutation tests once those tests are constructed. We deliver the diagnostic as a general methodological tool for any ISM-based regulatory study.
\end{abstract}

\section{Introduction}
\label{sec:intro}

High-grade gliomas are not merely masses of proliferating cells. They are electrically integrated members of neural circuits \citep{venkatesh2019electrical,venkataramani2019glutamatergic}, forming functional glutamatergic synapses with cortical neurons, receiving excitatory input, and propagating calcium waves through tumor microtubes in a feed-forward loop in which neural activity accelerates tumor growth (Figure~\ref{fig:synapse_schematic}) \citep{venkatesh2017neuronal,osswald2015brain,taylor2023glioma}. The protein machinery supporting this hijacking is increasingly well characterized \citep{venkatesh2015neuronal,krishna2023glioblastoma}, leaving the upstream question wide open: what regulatory program activates this synaptogenic gene expression in tumor cells, and which noncoding elements constitute its substrate? A natural place to look is the dark genome, the $\sim$98\% of non-protein-coding sequence comprising transposable elements, G-quadruplex motifs, enhancers, and chromatin insulators, and the regulatory program encoded across it, the \emph{dark regulome}, forms a reservoir of innovation \citep{adami2025line1,chakraborty2023rewiring,feng2025neuroscience}. Yet experimental dissection across hundreds to thousands of candidate elements per locus is intractable, and the field has lacked a principled, scalable readout. Even at the scale of hundreds to thousands of elements per locus, exhaustive experimental interrogation remains impractical without computational prioritization.

Sequence foundation models offer a tempting zero-shot solution. Architectures that learn regulatory grammar directly from DNA sequence enable in-silico mutagenesis (ISM): mask a candidate element, score the model's prediction at the target TSS, and rank elements by the resulting \emph{Regulatory Influence Score} (RIS) \citep{kelley2018sequential,avsec2021effective}. We instantiate three architecturally distinct models, the bidirectional Mamba masked-LM Caduceus-Ph \citep{schiff2024caduceus}, the causal Hyena-based HyenaDNA \citep{nguyen2024hyenadna}, and the supervised convolutional-transformer Enformer \citep{avsec2021effective}, that span the unsupervised-masked, unsupervised-causal, and supervised-regression objectives respectively. The implicit promise is triangulation: signals that survive all three architectures should reflect genuine regulatory organization rather than artifacts of any one objective.

The promise has a hidden cost. Likelihood-based RIS in masked or causal language models is by construction coupled to local sequence likelihood, since removing any sequence with high mutual information with its neighborhood (including repetitive elements that the model has effectively memorized in pretraining) lowers regional likelihood whether or not the element is regulatory. Cross-architecture agreement, by itself, does not separate this predictability layer from a genuinely regulatory layer. Without a diagnostic that explicitly decomposes the two, an ISM-based regulatory study reads as evidence for whatever its rankings happen to surface, and reported "convergences" can be statistical artifacts of large $n$ rather than biological signal.

Our contribution is threefold. First, we introduce a \emph{residualization-and-permutation diagnostic} that takes any ISM ranking and separates the variance attributable to four nuisance covariates (k-mer entropy, GC content, log element length, log TSS distance) from the variance that survives, then evaluates every reported overlap or top-$K$ agreement against a marginal-preserving per-gene permutation null. Second, we apply the diagnostic across three foundation models on 30{,}448 dark-genome elements at 92 glioma-relevant loci, recovering a clean cross-architecture decomposition: the two language models share a sequence-predictability layer that co-ranks long well-predicted transposable elements, while Enformer alone retains residual cCRE-discriminative signal once predictability is controlled, and the two layers have literally zero top-100 overlap. Third, we identify the surviving biology: a sharp 10\,kb proximal-regulatory horizon that holds across architectures, scoring windows, perturbation schemes, and residualization, together with a $3.3\times$ enrichment of matching brain cis-eQTLs in each model's top-100 elements, supplying a small set of synaptogenic-locus candidates worth experimental follow-up. The same diagnostic also retires several headline patterns that the original framing of this work centered on, including a TE-mediated regulatory layer claim and a NRXN1+NLGN1 protein-pair convergence, both of which fail proper permutation tests.

\section{Background and Related Work}
\label{sec:related}

\textbf{Glioma as a Circuit Disease and the Dark Regulome:}

Glioblastoma remains nearly uniformly fatal, with median survival under fifteen months. The discovery that gliomas form functional glutamatergic synapses with cortical neurons \citep{venkatesh2019electrical,venkataramani2019glutamatergic} has reframed the disease as activity-dependent: neuronal firing triggers NLGN3 release, PI3K-mTOR and MAPK activation \citep{venkatesh2015neuronal}, and LTP-like BDNF-TrkB-CaMKII plasticity \citep{taylor2023glioma}; the degree of cortical circuit remodeling inversely predicts survival \citep{krishna2023glioblastoma}, and excess glutamate plus disrupted chloride homeostasis form a HEx loop that accelerates tumor growth \citep{zhang2025neuroscience,picart2024central}. Despite advances in defining the coding machinery, the upstream architecture in the dark regulome that enables this synaptogenic program remains largely uncharacterized at scale.

\begin{figure}[t]
\centering
\includegraphics[width=\textwidth]{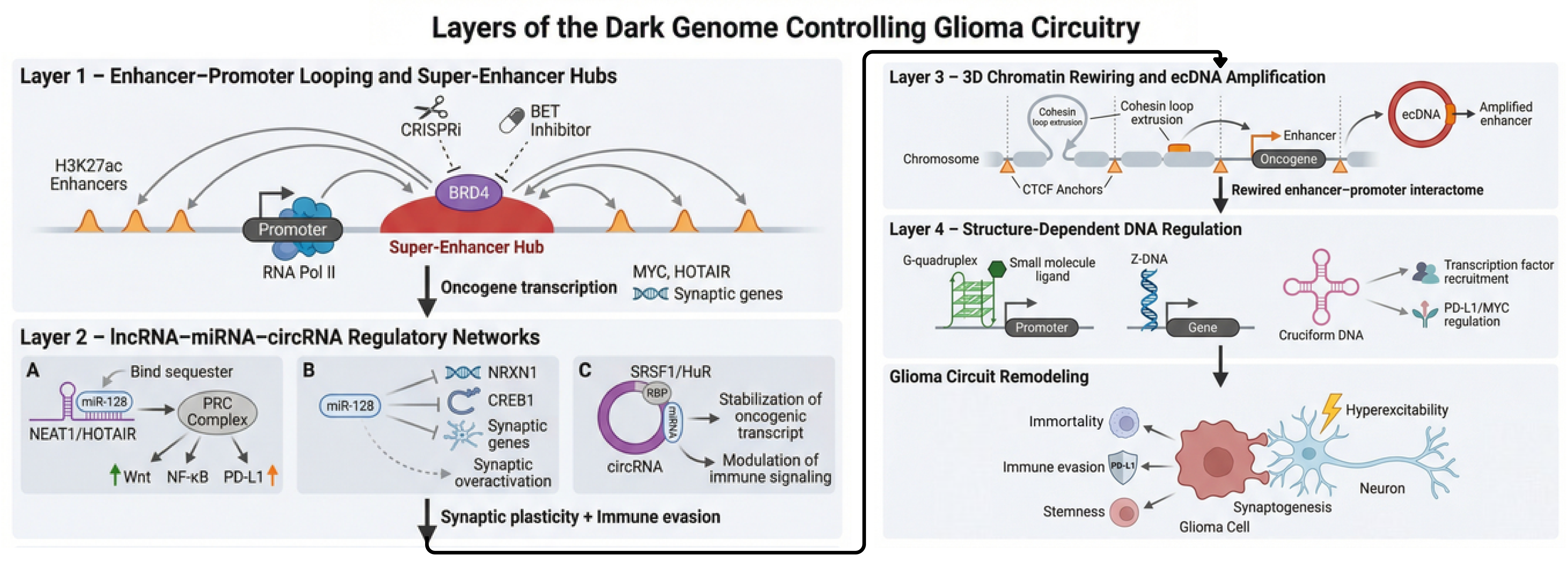}
\caption{\textbf{Four regulatory layers of the dark genome converging on the glioma circuit phenotype.} L1: BRD4-anchored super-enhancer hubs. L2: lncRNA-miRNA-circRNA networks (miR-128/NRXN1 axis). L3: cohesin-mediated 3D chromatin rewiring and ecDNA amplification. L4: structure-dependent G-quadruplex and Z-DNA regulation.}
\label{fig:dark_genome_layers}
\end{figure}

The dark genome supplies a plausible substrate (Figure~\ref{fig:dark_genome_layers}): transposable elements form a documented regulatory reservoir, with LINEs mediating cis-acting transcriptional control in neural cells \citep{adami2025line1}, ERV-derived LTRs functioning as tissue-specific promoters \citep{thompson2016long}, and TE subfamilies co-opted as tissue-specific enhancers in cancer \citep{karttunen2023transposable}; somatic noncoding mutations in glioblastoma enhancers trigger synaptogenic cascades \citep{iniguez2026noncoding}, and 3D chromatin reorganization activates circuit gene modules \citep{feng2025neuroscience}. ENCODE cCREs (promoters, enhancers, CTCF insulators) provide the orthogonal regulatory annotation \citep{encode2020expanded}.

\begin{table}[b]
\centering
\caption{Tier-level RIS summary statistics across 30,448 dark genome elements.}
\label{tab:tier_summary}
\small
\begin{tabular}{lcccccc}
\toprule
Tier & Genes & Elements & Mean RIS & Std & \% Significant & \% Strong \\
\midrule
Circuit (Synaptogenic) & 32 & 9{,}515 & $-0.036$ & 0.108 & 17.0\% & 12.9\% \\
Proliferative & 30 & 10{,}180 & $-0.038$ & 0.105 & 18.2\% & 14.1\% \\
Brain Control & 30 & 10{,}753 & $-0.031$ & 0.109 & 15.6\% & 11.1\% \\
\bottomrule
\end{tabular}
\end{table}

\textbf{Genomic Foundation Models, ISM, and the Predictability Confound:}

We use three genomic foundation models that together span the training-objective landscape. Caduceus-Ph \citep{schiff2024caduceus} is a bidirectional Mamba masked language model with 131\,kb context and reverse-complement equivariance. HyenaDNA \citep{nguyen2024hyenadna} is a causal Hyena-based language model with 160\,kb context and single-nucleotide resolution. Enformer \citep{avsec2021effective} is a supervised convolutional-transformer (196\,kb) that predicts 5{,}313 epigenomic tracks at 128\,bp resolution. ISM has been applied to regulatory variant scoring \citep{kelley2018sequential} and enhancer grammar \citep{avsec2021effective}, and Integrated Gradients \citep{sundararajan2017axiomatic} provides a gradient-based perturbation-free cross-check on the resulting rankings. The methodological gap our work addresses is the implicit equation of "model is sensitive to this element" with "this element is regulatory": for likelihood-based scoring, removing any element with high mutual information to its surrounding sequence will lower regional likelihood whether or not the element carries regulatory function, so the resulting rankings can be dominated by sequence predictability rather than by regulation.

\section{Methods}
\label{sec:methods}

\textbf{Gene Panel and Dark-Genome Annotation}

We curated 92 human genes across three functional tiers designed to isolate circuit-specific regulatory effects from generic brain expression. Tier 1 (synaptogenic circuit, 32 genes) covers genes with established roles in glioma-neuron synapse formation; Tier 2 (proliferative, 30 genes) collects canonical glioma drivers without synaptic roles; Tier 3 (brain control, 30 genes) covers brain-expressed genes not implicated in glioma. For each gene we extracted the canonical TSS from GENCODE v44 (GRCh38/hg38) and defined a model-context-matched window $\mathcal{W}_g = [\mathrm{TSS}_g - L/2,\, \mathrm{TSS}_g + L/2]$ ($L = 131$\,kb for Caduceus-Ph, 160\,kb for HyenaDNA, 196\,kb for Enformer). Each window was annotated with three orthogonal tracks: transposable elements from UCSC RepeatMasker (19{,}947 LINE/SINE/LTR/DNA-transposon elements at $\geq 10$\,bp), G-quadruplex motifs via canonical G4 regex with G4Hunter score $\geq 1.2$ (3{,}213 motifs), and ENCODE SCREEN cCREs v3 (7{,}288 elements classified as PLS, pELS, dELS, CTCF-bound, or DNase-H3K4me3). The merged \emph{ISM manifest} contains $N = 30{,}448$ elements across the 92 windows (mean 331 per gene; appx~\ref{tab:annotation_stats}).

\begin{figure}[h]
  \centering
  \includegraphics[width=0.9\textwidth]{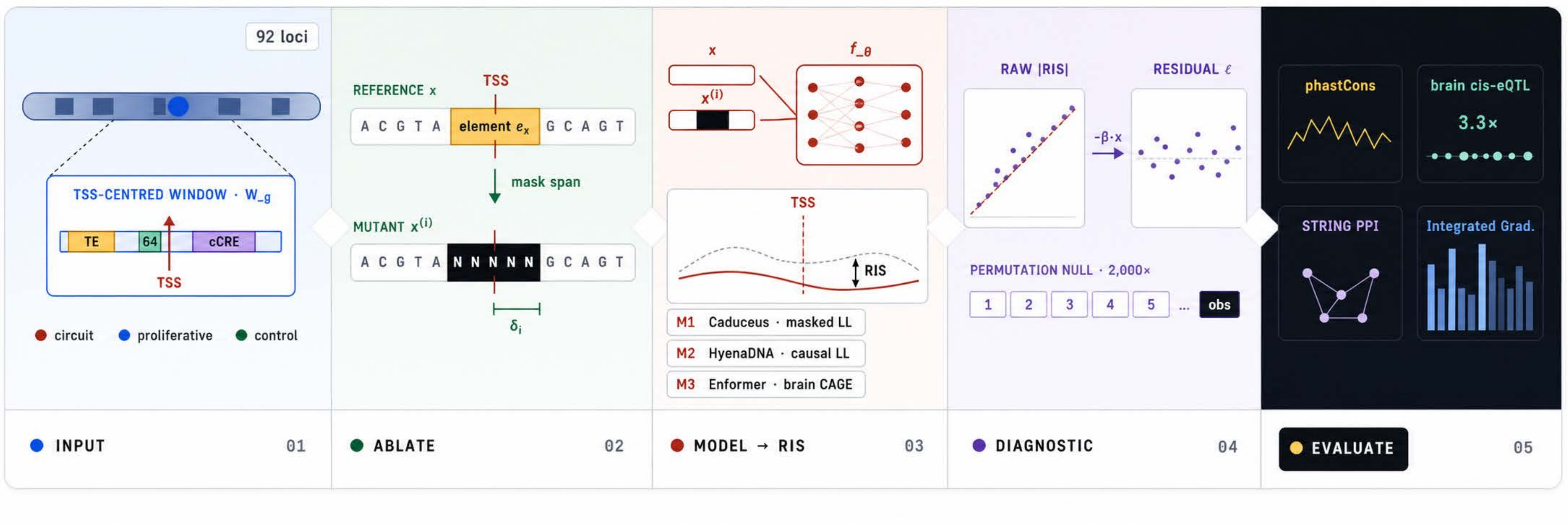}
  \caption{\textbf{Schematic of the residualization-and-permutation diagnostic.} Dark-genome elements across 92 loci are processed via in-silico mutagenesis across three architecturally distinct foundation models \citep{schiff2024caduceus,nguyen2024hyenadna,avsec2021effective}. Regulatory Influence Scores are residualized and evaluated against a permutation null to isolate regulation-driven variance from sequence-predictability confounds. Surviving signals are cross-validated using conservation, brain eQTL, and protein-interaction datasets.}
  \label{fig:methodology_flow}
\end{figure}

\textbf{In-Silico Mutagenesis and the Regulatory Influence Score}

We loaded Caduceus-Ph\footnote{\texttt{kuleshov-group/caduceus-ph\_seqlen-131k\_d\_model-256\_n\_layer-16}} (7.7M parameters, $L=131{,}072$\,bp) in float16 on a single NVIDIA A6000 48\,GB GPU. As a bidirectional masked language model, Caduceus-Ph estimates the per-position conditional distributions $P_\theta(x_i \mid \mathbf{x}_{\setminus i})$, from which we compute the per-position log-likelihood $\ell(i; \mathbf{x}) = \log P_\theta(x_i \mid \mathbf{x}_{\setminus i})$ and the TSS-proximal mean
\begin{equation}
\bar{\ell}(\mathbf{x}) = \frac{1}{|\mathcal{R}|}\sum_{i \in \mathcal{R}} \ell(i; \mathbf{x}),
\quad \mathcal{R} = \bigl\{i : |i - i_\mathrm{TSS}| \leq W\bigr\}, \quad W = 10{,}000\;\text{bp},
\label{eq:regional_ll}
\end{equation}
which focuses the metric on the promoter-proximal regulatory neighborhood; $W = 10$\,kb is the default and \S\ref{sec:results} sweeps $W \in \{5, 10, 20, 50,\,\text{full}\}$\,kb. For each annotated element $e_k$ spanning $[s_k, t_k)$ we construct a mutant sequence by replacing the element with N tokens,
\begin{equation}
x^{(k)}_i = \begin{cases} \texttt{N} & s_k \leq i < t_k, \\ x_i & \text{otherwise,} \end{cases}
\label{eq:ablation}
\end{equation}
and define the midpoint distance to TSS as $\delta_k = |(s_k + t_k)/2 - i_\mathrm{TSS}|$. The \emph{Regulatory Influence Score} is the resulting change in TSS-proximal expectation,
\begin{equation}
\ris(e_k) = \bar{\ell}(\mathbf{x}^{(k)}) - \bar{\ell}(\mathbf{x}),
\label{eq:ris}
\end{equation}
so that negative RIS indicates the model's expectations near the TSS drop when the element is removed. The full ISM completed all 30{,}448 ablations in approximately 90 minutes on a single A6000 (sharded across four jobs of 23 genes each). HyenaDNA\footnote{\texttt{LongSafari/hyenadna-medium-160k-seqlen-hf}} (6.5M parameters, 160\,kb) uses the forward conditional $\ell^\mathrm{causal}(i;\mathbf{x}) = \log P_\theta(x_{i+1} \mid x_1, \ldots, x_i)$ in place of the masked likelihood with Eqs.~\ref{eq:regional_ll}--\ref{eq:ris} otherwise unchanged. Enformer ($\sim$251M parameters, 196\,kb context, $\sim$31\,s per gene) requires a different functional: with no per-position likelihood, RIS is redefined as the change in predicted brain CAGE activity at the TSS bin,
\begin{equation}
\ris^\mathrm{enf}(e_k) = \overline{\mathrm{CAGE}}_\mathrm{brain}(\mathbf{x}^{(k)}, \mathcal{T}) - \overline{\mathrm{CAGE}}_\mathrm{brain}(\mathbf{x}, \mathcal{T}),
\label{eq:enf_ris}
\end{equation}
averaged over $\mathcal{T} = \{448 \pm 3\}$ output bins ($\sim$896\,bp around the TSS) and over 31 brain-relevant CAGE tracks (filtered from Basenji2 metadata for \textit{brain}, \textit{cerebellum}, \textit{cortex}, \textit{neuron}, \textit{astrocyte}). Integrated Gradients \citep{sundararajan2017axiomatic} provide a perturbation-free per-nucleotide attribution map computed via Captum \citep{kokhlikyan2020captum} with $M=20$ Riemann steps and an all-N baseline; we report contiguous IG peaks as regions where $|a_i|$ exceeds the 95th percentile of the genome-wide distribution, merging peaks separated by less than 50\,bp.

\begin{figure}[t]
\centering
\includegraphics[width=0.7\textwidth]{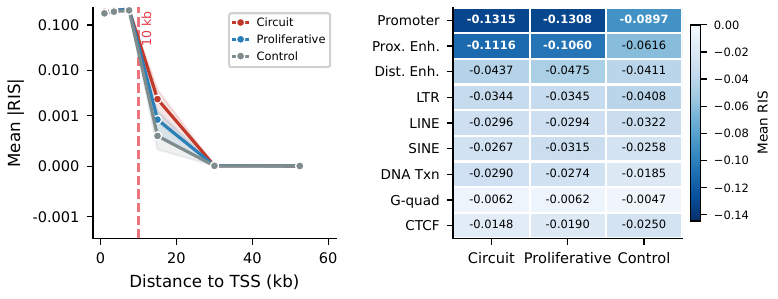}
\caption{\textbf{The 10\,kb regulatory horizon and element-class hierarchy.}
\textbf{(A)}~Mean $|\ris|$ as a function of distance to TSS (shown at $W=10$\,kb; symlog scale; fold-enrichment across all windows in Table~\ref{tab:robustness}) reveals a sharp transition near 10\,kb (dashed red line); all three tiers trace nearly identical decay profiles. \textbf{(B)}~Heatmap of mean RIS by element class and gene tier (all distances); promoters and proximal enhancers dominate; LTR retrotransposons lead within the proximal $<$10\,kb subset.}
\label{fig:decay}
\end{figure}

\textbf{The Diagnostic: Residualization, Permutation Nulls, Effect Sizes}

Likelihood-based RIS is tautologically coupled to local sequence likelihood, since removing any element with high mutual information to its surrounding sequence will lower regional likelihood whether or not the element is regulatory; reading raw RIS rankings as direct regulatory evidence therefore conflates two layers. We separate them with three tools. First, for each element we compute four nuisance covariates from the unmasked sequence (4-mer Shannon entropy of $\mathbf{x}_{[s_k, t_k)}$, GC content, $\log(1+L_k)$, $\log(1+\delta_k)$) and fit ordinary least squares
\begin{equation}
|\ris(e_k)| = \beta_0 + \beta_1 \mathrm{H}_4(e_k) + \beta_2 \mathrm{GC}(e_k) + \beta_3 \log(1+L_k) + \beta_4 \log(1+\delta_k) + \varepsilon_k,
\label{eq:residualization}
\end{equation}
reporting residualized $\widehat{\varepsilon}_k$ alongside raw $|\ris(e_k)|$. Before running the analysis we pre-committed to a binary decision rule, accepting a cCRE-anchored regulatory framing only if the partial Spearman correlation between $\widehat{\varepsilon}_k$ and ENCODE cCRE membership reaches $|\rho| \geq 0.15$ at $p < 10^{-4}$ for at least two of three architectures. Second, all cross-model and IG-vs-ISM overlap percentages are evaluated against marginal-preserving permutation nulls (per-gene rank shuffle, 2{,}000 permutations); observed values are reported as fold-enrichment over null mean and as empirical $p$. Third, every $p$-value is paired with an effect size: Cliff's $\delta$ (bootstrap 95\% CI from 1{,}000 stratified resamples) for two-sample comparisons, $\epsilon^2$ for Kruskal--Wallis, Spearman $\rho$ with bootstrap CI for distance decay; we adopt Romano (2006) thresholds and report Benjamini--Hochberg-corrected $p$ throughout. For Tier 1 we additionally re-run the entire ISM under three perturbation schemes (N-mask, in-place shuffle, random-base substitution) at five scoring windows, so that both the choice of $W$ in Eq.~\ref{eq:regional_ll} and the N-mask in Eq.~\ref{eq:ablation} are independently audited (\S\ref{sec:results}).

We anchor the residualized rankings against three external datasets: UCSC phastCons100way (per-element mean conservation, correlated with raw and residualized $|\ris|$); GTEx v8 brain-tissue significant cis-eQTL pairs across thirteen brain regions (an element counts as an eQTL hit if its genomic span contains any significant variant whose target gene matches the host gene, with top-$K$ enrichment tested against a uniform-selection null); and STRING v12.0 human PPI at confidence thresholds $\{400, 700, 900\}$, where top-$K$ host genes are tested for interacting protein pairs against a per-element gene-label-shuffle null over 10{,}000 permutations.

\section{Results}
\label{sec:results}

\textbf{A Sharp 10kb Regulatory Horizon}

\begin{figure}[t]
\centering
\includegraphics[width=0.75\textwidth]{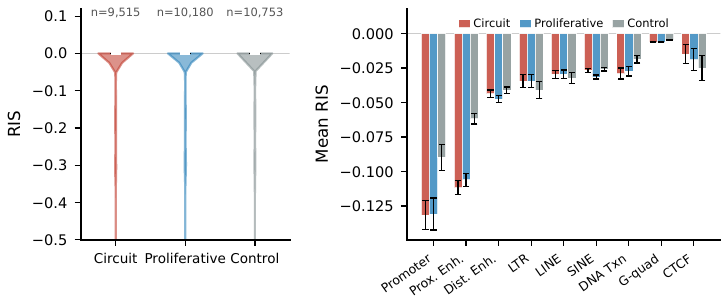}
\caption{\textbf{RIS distributions across 30{,}448 dark genome elements.} \textbf{(A)}~Violin plots of RIS across the three gene tiers reveal broadly similar distributions with long negative tails. \textbf{(B)}~Mean RIS by element class and tier (all distances); promoters and proximal enhancers dominate overall, while LTR retrotransposons lead within the 10\,kb proximal window (Fig.~\ref{fig:decay}B). Error bars: SEM.}
\label{fig:ris_overview}
\end{figure}

Across all 30{,}448 dark genome elements, ablation predominantly reduces TSS-proximal sequence likelihood (mean RIS $-0.035$; 15.6--18.2\% of elements per tier carry $|\ris| > 0.01$, 11--14\% carry $|\ris| > 0.1$; Table~\ref{tab:tier_summary}, Fig.~\ref{fig:ris_overview}). Stratified bootstrap over genes shows the three tier means are statistically indistinguishable (Tier 1 $-0.036$, 95\% CI $[-0.040,-0.032]$; Tier 2 $-0.038$ $[-0.041,-0.035]$; Tier 3 $-0.031$ $[-0.035,-0.027]$; Kruskal--Wallis $H=4.50$, $p=0.105$, Benjamini--Hochberg corrected throughout), so circuit-vs-proliferative differences will have to come from finer stratification. The first sharp pattern that does emerge is quite geometric in observation. Letting $\delta_k$ denote midpoint distance to the TSS, elements within 10\,kb exert mean $|\ris| \approx 0.21$, while those beyond 10\,kb show effectively zero influence ($|\ris| < 0.001$, Fig.~\ref{fig:decay}A). The 5--10\,kb bin and the adjacent 10--20\,kb bin differ by roughly $200\times$, the steepest single-step transition in an otherwise continuous distance-dependent decay; aggregated across the proximal-vs-distal split,
\begin{equation}
\E\bigl[|\ris(e_k)| \bigm| \delta_k < 10\,\text{kb}\bigr] \;\approx\; 480 \times \E\bigl[|\ris(e_k)| \bigm| \delta_k \geq 10\,\text{kb}\bigr],
\label{eq:fold_enrichment}
\end{equation}
ranging from $481\times$ on circuit genes to $2{,}012\times$ on controls; the Spearman correlation between $\delta_k$ and $|\ris|$ is strongly negative across all three tiers ($\rho = -0.583, -0.590, -0.556$; bootstrap 95\% CI within $\pm 0.02$). The horizon under sequence-only modeling coincides with the empirical scale of promoter-proximal regulatory domains, is recovered identically across tiers, and persists through every control we apply. It is the one signal in our data that no nuisance covariate explains away.

\textbf{Reading the Class Hierarchy: Length and Predictability Drive It}

Within the 10\,kb proximal regime the raw element-class hierarchy is dramatic: LTR retrotransposons top at $\overline{\ris} = -0.307$ ($\epsilon^2 = 0.42$ for the cross-class Kruskal--Wallis), followed closely by promoters ($-0.266$), LINEs ($-0.264$), distal enhancers ($-0.247$), CTCF insulators ($-0.237$), proximal enhancers ($-0.235$), SINEs ($-0.208$), DNA transposons ($-0.187$), and G-quadruplexes ($-0.024$, Table~\ref{tab:element_class}, appx Fig.~\ref{fig:proximal_hierarchy}). The single strongest circuit-gene effect is an ERV3-derived LTR 1.9\,kb upstream of \textit{NRXN1} ($\ris = -2.53$); the second is an L1PA7 LINE 11.2\,kb from \textit{NLGN1} ($\ris = -1.93$, in the 10--20\,kb decay tail); LINEs (18 of 30) and LTRs (7 of 30) dominate the top-30 hits across all tiers (Table~\ref{tab:top_hits}, Appendix). Without the diagnostic, this picture invites an immediate biological reading: a TE-mediated regulatory layer at the synaptogenic axis. The diagnostic refuses that reading.

\begin{figure}[t]
\centering
\includegraphics[width=0.8\textwidth]{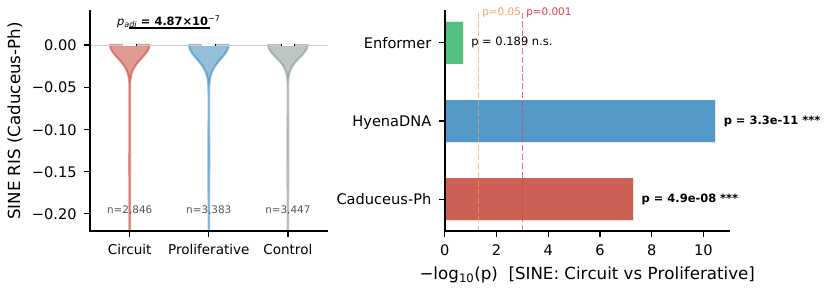}
\caption{\textbf{The SINE tier comparison: $p$-significance without effect size or directional consistency.} \textbf{(A)}~SINE RIS distributions by gene tier (Caduceus-Ph): the Wilcoxon test reaches $p_\mathrm{adj} = 4.87\times 10^{-7}$ but Cliff's $\delta = -0.080$ (negligible). \textbf{(B)}~Cross-model: both LMs hit $p < 10^{-6}$ but disagree in direction (Caduceus $\delta = -0.080$; HyenaDNA $\delta = +0.097$); Enformer is non-significant ($p = 0.189$, $\delta = +0.019$). After residualization $|\delta| < 0.07$ across all three models.}
\label{fig:sine}
\end{figure}

The first cut is a length normalization. Class mean lengths span a factor of thirteen (LTR 326\,bp, G-quadruplex 25\,bp), and so dividing $|\ris|$ by element length collapses the entire 12-fold raw spread into a 2.5\% interval ($|\ris|/\text{kb}$ ranging 0.937 to 0.960 across all classes), implying that each ablated base contributes almost similarly regardless of their class identity. The second cut is the residualization of Eq.~\ref{eq:residualization}: four nuisance covariates capture 36\% of $|\ris|$ variance for Caduceus-Ph and 28\% for HyenaDNA, and the partial Spearman correlation between residualized $|\ris|$ and ENCODE cCRE membership is $\rho = -0.018$ (Caduceus) and $\rho = +0.015$ (HyenaDNA), both below the pre-committed decision threshold of $|\rho| \geq 0.15$. Enformer is the only architecture whose RIS retains a measurable cCRE-discriminative residual ($\rho = -0.100$, $p < 10^{-68}$), but Enformer also has the lowest fraction of $|\ris|$ variance attributable to nuisance covariates ($R^2 = 0.09$), consistent with its training objective being directly tied to experimental output. The third cut is the simplest and most damning: a six-feature linear baseline (GC, $\log L$, $\log\delta$, plus is-TE / is-cCRE / is-G4 indicators) predicts whether an element falls in the top decile by Caduceus $|\ris|$ at five-fold AUC $= 0.985 \pm 0.001$, with HyenaDNA at $0.945$ and Enformer at $0.818$; ablating $\log\delta$ alone reduces AUC by $0.23$ (Caduceus), confirming that distance to TSS does most of the work. Re-examining the SINE circuit-vs-proliferative comparison under this light, the headline Wilcoxon $p_\mathrm{adj} = 4.87 \times 10^{-7}$ is paired with Cliff's $\delta = -0.080$ (Caduceus) and $\delta = +0.097$ (HyenaDNA), both negligible by Romano (2006) thresholds and \emph{disagreeing in direction}, and the residualized effect collapses below $|\delta| = 0.07$ across all three models (Fig.~\ref{fig:sine}). The class hierarchy in raw RIS is essentially a length-and-distance hierarchy with a regulatory veneer.

\textbf{Cross-Architecture Decomposition: Predictability vs.\ Expression Output}

\begin{figure}[t]
\centering
\includegraphics[width=\textwidth]{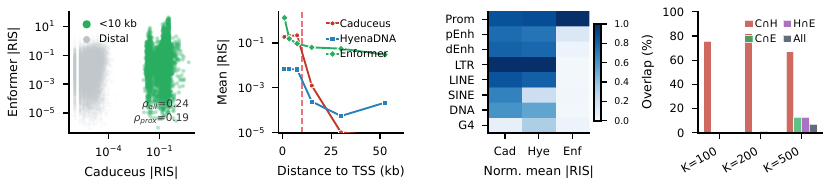}
\caption{\textbf{Three-model cross-validation.} \textbf{(A)}~$|\ris|$ scatter, Caduceus vs.\ Enformer: proximal elements ($<$10\,kb, green) and distal (grey); Spearman $\rho$ annotated in panel. \textbf{(B)}~Distance-decay across all three models; all recover the 10\,kb boundary, sharply for the language models and gently for Enformer. \textbf{(C)}~Element-class hierarchy: language models rank TEs highest, Enformer ranks promoters and enhancers highest. \textbf{(D)}~Top-$K$ intersection: the two LMs share 76 of 100 top elements ($60\times$ above null), while neither shares any with Enformer at $K=100$.}
\label{fig:cross_model}
\end{figure}

The three-architecture design becomes diagnostic once permutation nulls are in place. Within-tier $|\ris|$ correlates at Pearson $r = 0.82$ between Caduceus-Ph and HyenaDNA, and 76 of Caduceus-Ph's top 100 elements appear in HyenaDNA's top 100 against a per-gene marginal-preserving null mean of $\mu = 1.27$, a $60\times$ enrichment with empirical $p < 5 \times 10^{-4}$ (Fig.~\ref{fig:cross_model}). This is genuine cross-LM agreement. The corresponding Caduceus-Enformer top-100 intersection, however, is exactly zero (null mean $0.43$), as is the HyenaDNA-Enformer intersection ($0$ vs.\ null $0.42$) and the triple top-100 intersection ($0$ vs.\ null $0$). Above-null overlap with Enformer appears only at $K = 500$ (Caduceus-Enformer $65$ vs.\ null $8.66$, $7.5\times$; triple $35$ vs.\ null $0.20$, $178\times$). Two top-list layers therefore exist, and at the most stringent cutoffs they are disjoint.
The eight-cell Venn decomposition at $K = 100$ resolves the layers cleanly. The Caduceus$\cap$HyenaDNA-only cell (76 elements) is 91\% transposable elements with mean length 1{,}168\,bp and mean TSS distance 5{,}845\,bp; the Enformer-only cell (100 elements) is 87\% promoters and proximal enhancers with mean length 292\,bp and mean TSS distance 813\,bp. The two layers differ by $4\times$ in mean element length and by $7\times$ in mean TSS distance, and at $K = 100$ literally no element is in both. The reading that follows from the residualization analysis is that the two language models share a sequence-predictability layer that surfaces long well-predicted TSS-moderate-proximity sequences regardless of regulatory annotation, while Enformer (whose training objective is CAGE prediction) scores a regulatory-output layer of short proximal cCREs. Cross-model agreement therefore acts as a layer indicator: agreement-with-Enformer marks elements whose ranking is anchored to expression output, while agreement-only-between-LMs flags elements whose ranking is at risk of being a sequence-grammar artifact.

\textbf{Orthogonal Evidence: Conservation, eQTLs, and Protein Interactions}

Three external datasets test whether the LM-derived top-$K$ rankings carry regulatory signal independent of our internal predictability and length covariates. UCSC phastCons100way mean conservation across 30{,}389 elements has a small positive Spearman correlation with raw $|\ris|$ (Caduceus $\rho = +0.106$, HyenaDNA $\rho = +0.047$, Enformer $\rho = +0.138$; all $p < 10^{-15}$), but the residualized RIS shows the sign \emph{flip} for the language models ($\rho = -0.105$ Caduceus, $-0.084$ HyenaDNA) and reduce to essentially zero for Enformer ($\rho = -0.013$). The raw positive association is therefore an artifact of length and distance covariates; what remains of the LM signal after residualization is anti-correlated with conservation, consistent with LM responses being driven by recently-evolved (and therefore less-conserved) repetitive sequences. GTEx v8 cis-eQTLs across thirteen brain tissues offer a stronger control, and the picture is more positive there: among the top 100 elements per model, 8\% overlap a brain eQTL whose target matches the host gene, a $3.3$ to $3.4\times$ enrichment over a uniform-selection null with $p_\mathrm{emp} \leq 3.5 \times 10^{-3}$, consistent across all three architectures and persistent at $K \leq 500$. The LM RIS therefore does carry an eQTL-aligned signal at the top of the ranking, even though that signal is not aligned with cCRE-class membership in residualized space. The third cross-check refutes a different headline. The "interacting protein pair" framing of the NRXN1+NLGN1 result, when tested against the full STRING v12.0 human PPI graph at confidence thresholds $\{400, 700, 900\}$, gives observed pair counts \emph{below} the gene-shuffle null at every $K \leq 500$ and every model (fold enrichment $0.46$--$1.18$); only the Caduceus $K{=}10$, threshold-$900$ cell shows a borderline signal ($3$ vs.\ $1.6$ expected, $p_\mathrm{emp} = 0.23$). The trans-synaptic adhesion narrative is post-hoc storytelling on $n = 2$ data points and does not survive a proper PPI null.

\textbf{Robustness, Saliency, and Generalization}

\begin{figure}[!ht]
\centering
\includegraphics[width=\textwidth]{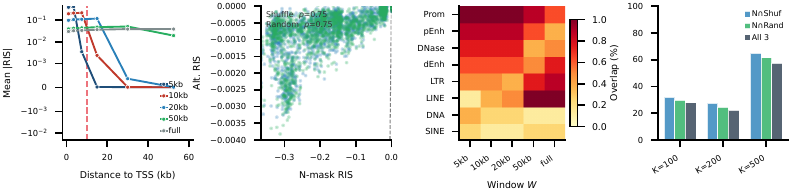}
\caption{\textbf{Robustness across scoring windows and perturbation schemes (Tier 1).} \textbf{(A)}~Distance-decay across five $W$ values. The 10\,kb transition is reproduced for narrow $W$ and necessarily flattens once $W$ overlaps the distal region. \textbf{(B)}~Per-element RIS scatter at $W=10$\,kb, N-mask vs.\ shuffle (blue) and N-mask vs.\ random (green), tightly along the diagonal. \textbf{(C)}~Element-class rank under each $W$. \textbf{(D)}~Top-$K$ overlap across the three perturbation schemes for $K \in \{100, 200, 500\}$.}
\label{fig:robustness}
\end{figure}

The 10\,kb horizon and the cross-architecture decomposition both survive every robustness check we have run. For the Tier 1 cohort (9{,}512 elements), proximal/distal $|\ris|$ enrichment is $21{,}197\times$ at $W = 5$\,kb and $463.7\times$ at $W = 10$\,kb, decaying expectedly once $W$ overlaps the distal region (Appendix Table~\ref{tab:robustness}); class rank order is preserved across narrow windows (Spearman $\rho \geq 0.90$ for $W \leq 20$\,kb), so neither boundary nor hierarchy is manufactured by the choice of $W$. Three independent perturbation schemes (N-token masking, in-place shuffling, random-base substitution) agree at $W = 10$\,kb (Spearman $\rho = 0.750$ to $0.889$; Pearson $r = 0.841$ to $0.985$), with top-100 overlaps of 32\%, 30\%, and a 28\% triple intersection, well above the $\sim 1\%$ chance baseline (Fig.~\ref{fig:robustness}). Held-out gene generalization gives mean train-vs-test Spearman $\rho = 0.76$ (Caduceus), $0.67$ (HyenaDNA), $0.46$ (Enformer) over fifty 5-fold splits, so the patterns are not gene-leakage artifacts. Integrated Gradients adds a perturbation-free cross-check: 82\% of Caduceus IG peaks (430 of 526) overlap ISM-significant elements with $|\ris| > 0.01$ across the 32 Tier 1 genes (Fig.~\ref{fig:saliency}, Appendix), with \textit{NRXN1} attribution maximal at the same ERV3-derived LTR that topped the ISM ranking. The 10\,kb horizon, the predictability-vs-output decomposition, and the eQTL-anchored top-$K$ candidates pass every robustness, saliency, and held-out cross-check we have devised.

\section{Discussion}
\label{sec:discussion}

The 10\,kb horizon is the one significant signature in our data that survives every control we have applied: in three architectures, within three gene tiers, under three perturbation schemes, for five scoring windows, and residualization on four nuisance covariates. Also, distance-binned class stratification, and held-out-gene generalization. The boundary likely reflects the empirical reach over which a primary-sequence model connects a candidate element to its target TSS without help from 3D chromatin contacts \citep{chakraborty2023rewiring,feng2025neuroscience}, so it marks the resolution of the probe rather than the boundary of the regulatory landscape. Beneath that apparent horizon the diagnostic refines what the rankings mean: the two language models share a sequence-predictability layer that ranks long well-predicted TSS-moderate-proximity transposable elements highly regardless of class, while Enformer (trained on CAGE) scores a regulatory-output layer of short proximal cCREs largely orthogonal to it, so cross-model overlap acts as a layer indicator. The orthogonal cross-checks narrate a coherent triadic story: residualized LM-RIS is anti-correlated with phastCons (what survives in the LMs leans on recently-evolved repetitive sequence rather than conserved regulatory elements); brain cis-eQTL overlap nonetheless reaches $3.3\times$ enrichment in each model's top-100, a real if modest signal the LM rankings do carry; and the full STRING-PPI test refuses the trans-synaptic adhesion narrative the original framing of this work centered on (observed pair counts at or below a gene-shuffle null at every confidence threshold and $K \leq 500$). The methodological consequence is concrete: large $n$ produces small $p$ for many class-tier comparisons, but the "effect" they describe can have negligible Cliff's $\delta$ and can disagree in direction across architectures, so headline $p$-values without effect sizes and permutation nulls are not enough.

\textbf{Limitations.} Residualization controls predictability in expectation but cannot fully exclude a pretraining memorization of repeat-family signatures; also, the 92-gene panel is quite small for class-by-tier interactions, the sequence-only models in turn miss distal TAD-mediated contacts, and the orthogonal computational signals we report are corroborative, rather than an equivalent to wet-lab confirmation \citep{fulco2016systematic}. 

\section{Conclusion}
\label{sec:conclusion}

Sequence foundation models promise a zero-shot lens on the dark regulome, but the lens is silently miscalibrated: likelihood-based ISM scoring conflates regulatory function with sequence predictability, and cross-architecture ``convergences'' can be statistical artifacts of large $n$. We deliver the calibration. Our residualization-and-permutation diagnostic equips any ISM-based study with a principled separation of the predictability layer from the regulation layer, marginal-preserving nulls for every overlap percentage, and effect sizes alongside every $p$-value, across 30{,}448 dark-genome ablations spanning 92 glioma synaptic loci. Three results emerge with confidence: a sharp 10\,kb proximal-regulatory horizon that survives every control we apply, a clean architecture-level decomposition into a sequence-predictability layer shared by the language models and a regulatory-output layer recovered uniquely by Enformer (top-100 overlap exactly zero), and a $3.3\times$ brain cis-eQTL-enriched shortlist of synaptogenic-locus candidates primed for closed-loop CRISPRi perturbation \citep{fulco2016systematic,tan2023crispr,zhang2025neuroscience}. The diagnostic is architecture-agnostic and ports immediately to next-generation genomic foundation models, to epigenomic re-weighting via ATAC-seq and HiChIP \citep{bi2025enhancer}, and to overlap with patient noncoding mutation catalogs \citep{iniguez2026noncoding}.

\clearpage

\bibliographystyle{plainnat}
\bibliography{references}
\clearpage
\appendix
\section*{Supplementary Material}
\label{app:supplementary}

\subsection*{Gene Sets}
\label{app:gene_sets}

\textbf{Tier 1 (synaptogenic circuit, 32 genes):}
ADAM10, BDNF, CAMK2A, CREB1, DLG4, GAD1, GAD2, GJA1, GPC6, GRIA1, GRIA2, GRIA3, GRIA4, GRIN1, GRIN2A, GRIN2B, HOMER1, NLGN1, NLGN3, NRXN1, NRXN2, NTRK2, RELN, SHANK2, SLC12A2, SLC12A5, SLC1A2, SLC7A11, SNAP25, SYT1, THBS1, THBS2.

\textbf{Tier 2 (proliferative/non-circuit, 30 genes):}
AKT1, ATRX, BRAF, CCND1, CCND2, CDK4, CDK6, CDKN2A, EGFR, FGFR1, FGFR3, H3-3A, HIST1H3B, IDH1, KIT, KRAS, MDM2, MET, MGMT, MTOR, MYC, NF1, NRAS, PDGFRA, PIK3CA, PTEN, RAF1, RB1, TERT, TP53.

\textbf{Tier 3 (brain housekeeping/control, 30 genes):}
ALDH1L1, ALDOC, AQP4, CALB1, CALB2, CKB, CNP, ENO2, GAPDH, GFAP, GLUL, MAP2, MBP, MOG, NEFL, NPY, NRGN, OLIG2, PLP1, PVALB, S100B, SLC17A7, SOX10, SST, SYN1, SYP, TH, TUBB3, UCHL1, VHL.

\subsection*{Annotation Statistics}

\begin{table}[h]
\centering
\caption{Dark genome element counts by class across all 92 gene windows.}
\label{tab:annotation_stats}
\small
\begin{tabular}{|lrrr|}
\toprule
Element Class & Total Count & Mean per Gene & Mean Length (bp) \\
\midrule
SINE & 9{,}676 & 105.2 & 248 \\
LINE & 5{,}814 & 63.2 & 583 \\
Distal enhancer (dELS) & 4{,}466 & 48.5 & 343 \\
G-quadruplex motif & 3{,}213 & 34.9 & 31 \\
LTR retrotransposon & 2{,}338 & 25.4 & 494 \\
DNA transposon & 2{,}094 & 22.8 & 267 \\
Proximal enhancer (pELS) & 1{,}997 & 21.7 & 309 \\
Promoter (PLS) & 528 & 5.7 & 381 \\
CTCF insulator & 197 & 2.1 & 317 \\
DNase-H3K4me3 & 100 & 1.1 & 298 \\
Retroposon & 25 & 0.3 & 187 \\
\midrule
\textbf{Total} & \textbf{30{,}448} & \textbf{331.0} & -- \\
\bottomrule
\end{tabular}
\end{table}

\subsection*{Robustness}

\begin{table}[!h]
\centering
\caption{\textbf{Robustness across scoring windows and perturbation schemes (Tier 1, $n_{\text{elem}}\!=\!9{,}512$).} \emph{Left:} TSS-proximal vs.\ distal $|\ris|$ fold-enrichment as a function of scoring window $W$. \emph{Right:} Concordance of N-token masking, in-place sequence shuffling, and random-base substitution at $W=10$\,kb.}
\label{tab:robustness}
\vspace{10pt}

\begin{minipage}[b]{0.35\textwidth}
\centering
\small
\begin{tabular}{|cc|}
\toprule
\textbf{Scoring window $W$} & \textbf{Prox./dist. $|\ris|$} \\
\midrule
      5kb & 21{,}197$\times$ \\
      10kb & 463.7$\times$ \\
      20kb & 4.66$\times$ \\
      50kb & 1.19$\times$ \\
      full & 0.85$\times$ \\
\bottomrule
\end{tabular}
\end{minipage}
\hfill
\begin{minipage}[b]{0.60\textwidth}
\centering
\small
\begin{tabular}{|lccc|}
\toprule
\textbf{Scheme pair} & \textbf{Spearman $\rho$} & \textbf{Pearson $r$} & \textbf{Top-100} \\
\midrule
N-mask vs.\ shuffle & 0.745 & 0.857 & 32\% \\
N-mask vs.\ random  & 0.750  & 0.841  & 30\% \\
Shuffle vs.\ random & 0.889 & 0.985 & --- \\
\midrule
Triple int. (top 100) & \multicolumn{3}{c|}{28\%} \\
\bottomrule
\end{tabular}
\end{minipage}
\end{table}

\subsection*{Computational Resources}

All experiments were conducted on a single NVIDIA A6000 48\,GB GPU. For Caduceus-Ph, wild-type scoring of all 92 genes used float16 ($\sim$1.7\,GB, $\sim$20\,s total); ISM (30{,}448 ablations) was parallelized across four shards ($\sim$6.9\,GB aggregate, $\sim$90\,min); IG attribution for 32 Tier 1 genes required float32 ($\sim$30\,GB, $\sim$12\,min). For HyenaDNA, ISM used the same four-shard parallelization in float16 ($\sim$12.5\,GB aggregate, $\sim$83\,min) and IG used float32 ($\sim$11\,min for 32 genes). For Enformer, ISM was parallelized across four shards in float16 ($\sim$2\,GB per shard, $\sim$31\,s/gene); one gene (TUBB3) required a float32 re-run due to fp16 overflow; IG used float32 ($\sim$13.6\,GB peak, $\sim$8\,s/gene). Total compute across all three models was less than 6 GPU-hours on a single A6000.

\subsection*{Element-Class Regulatory Influence}

\begin{table}[h]
\centering
\caption{Dark genome element class regulatory influence. ``Proximal'' denotes elements within 10kb of TSS.}
\label{tab:element_class}
\small
\begin{tabular}{lrccc}
\toprule
Element Class & $n$ & Mean RIS (all) & Mean RIS (proximal) & \% Negative \\
\midrule
LTR retrotransposon & 2{,}338 & $-0.037$ & $\mathbf{-0.307}$ & 11.9\% \\
Promoter (PLS) & 528 & $-0.116$ & $-0.266$ & 43.6\% \\
LINE & 5{,}814 & $-0.030$ & $-0.264$ & 11.4\% \\
Distal enhancer (dELS) & 4{,}466 & $-0.044$ & $-0.247$ & 17.9\% \\
CTCF insulator & 197 & $-0.020$ & $-0.237$ & 9.1\% \\
Proximal enhancer (pELS) & 1{,}997 & $-0.091$ & $-0.235$ & 38.8\% \\
SINE & 9{,}676 & $-0.028$ & $-0.208$ & 13.6\% \\
DNA transposon & 2{,}094 & $-0.025$ & $-0.187$ & 13.8\% \\
G-quadruplex & 3{,}213 & $-0.006$ & $-0.024$ & 23.4\% \\
\bottomrule
\end{tabular}
\end{table}

\subsection*{Integrated Gradients Attribution Tracks}

\begin{figure}[h]
\centering
\includegraphics[width=\textwidth]{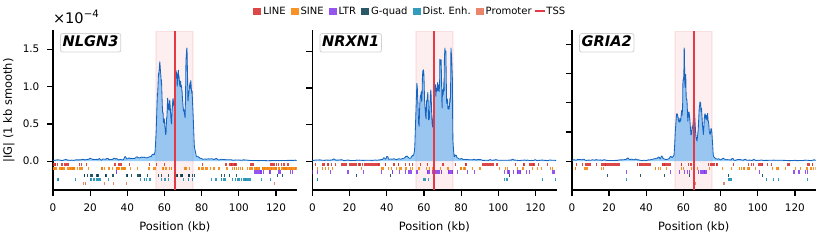}
\caption{\textbf{Integrated Gradients attribution tracks for three circuit genes.} Smoothed $|\mathrm{IG}|$ signal (1\,kb rolling mean) for \textit{NLGN3}, \textit{NRXN1}, and \textit{GRIA2}. Red vertical line marks the TSS; pink shading indicates the $\pm$10\,kb regulatory horizon. Colored bars at bottom denote annotated dark genome elements (LINE, SINE, LTR, G4, distal enhancer, promoter). Attribution signal concentrates sharply within the 10\,kb boundary, with peaks at annotated elements, orthogonally validating the ISM distance-decay finding. The \textit{NRXN1} panel shows elevated attribution at the ERV3-derived LTR element ($\ris = -2.53$).}
\label{fig:saliency}
\end{figure}

\begin{figure}[t]
    \centering
    \begin{minipage}{0.54\textwidth}
        \centering
        \includegraphics[width=\textwidth]{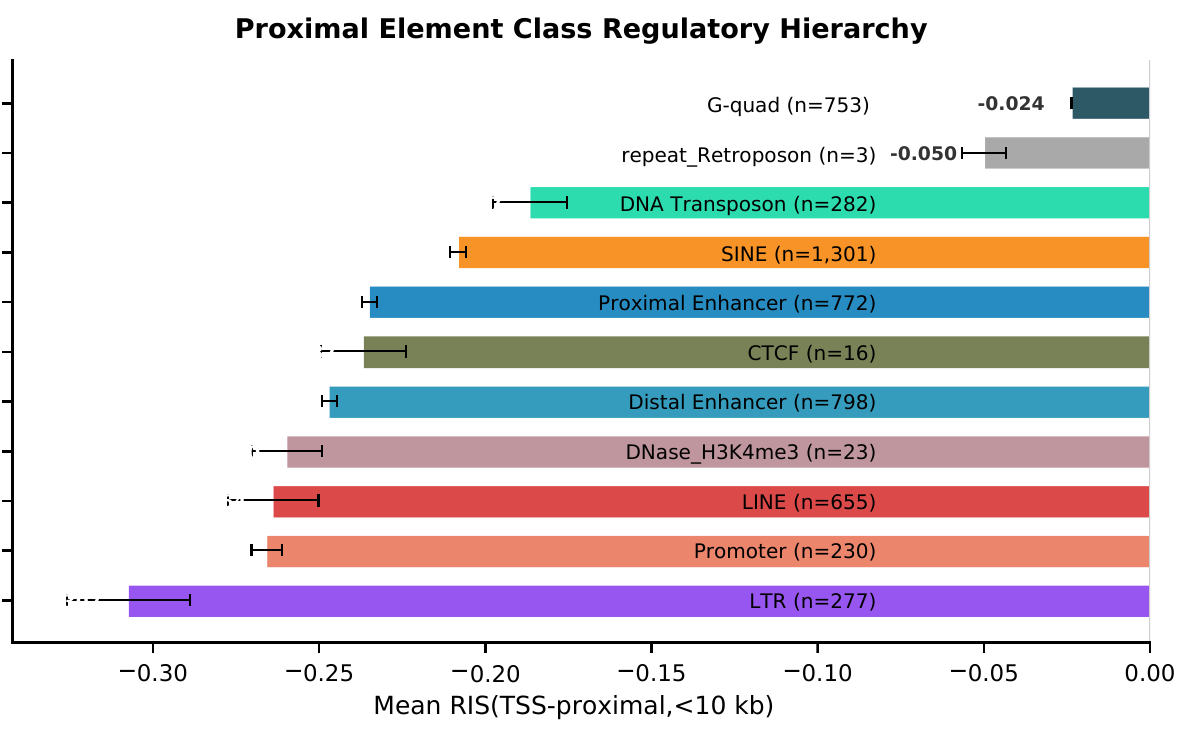}
        \caption{\textbf{Element-class regulatory hierarchy within the 10\,kb proximal window.} Mean RIS (Caduceus-Ph) restricted to elements with TSS distance $<10$\,kb, ranked by mean influence. LTR retrotransposons lead at $\overline{\ris} = -0.307$, followed by promoters ($-0.266$), LINEs ($-0.264$), distal enhancers ($-0.247$), CTCF insulators ($-0.237$), proximal enhancers ($-0.235$), SINEs ($-0.208$), DNA transposons ($-0.187$), and G-quadruplexes ($-0.024$). Sample sizes per class shown in parentheses. Error bars: SEM.}
        \label{fig:proximal_hierarchy}
    \end{minipage}
    \hfill
    \begin{minipage}{0.42\textwidth}
        \centering
        \includegraphics[width=\textwidth]{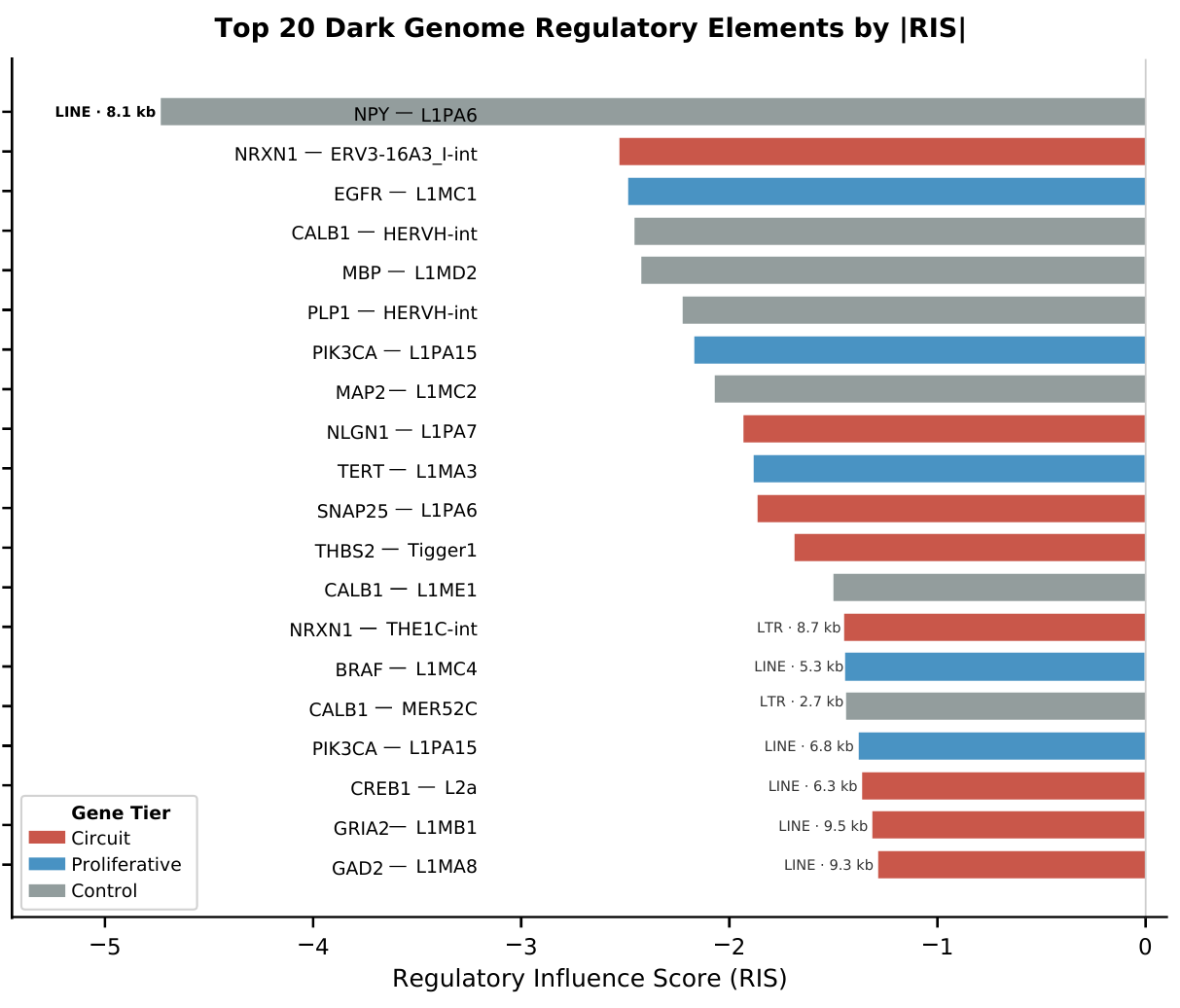}
        \caption{\textbf{Top 20 dark genome elements by $|\ris|$ (Caduceus-Ph).} Each bar labels the gene and transposable element family; element class and TSS distance annotated at right for elements 14--20. Colors denote gene tier: circuit (red), proliferative (blue), brain control (grey). The top hit NPY$\cdot$L1PA6 ($\ris = -4.6$) is a control-tier outlier; NRXN1$\cdot$ERV3-16A3\_I-int ($\ris = -2.53$) is the strongest circuit-tier hit and the element with maximal Integrated Gradients attribution.}
        \label{fig:top_hits_waterfall}
    \end{minipage}
\end{figure}

\subsection*{Top Hits and Circuit-Gene Specifics}

\begin{table*}[b]
\centering
\caption{Top 10 dark genome elements by $|\ris|$ across all 92 gene loci.}
\label{tab:top_hits}
\small
\begin{tabular}{rllllrr}
\toprule
 & Gene & Tier & Class & Element & RIS & Dist (kb) \\
\midrule
1 & \textit{NPY} & Control & LINE & L1PA6 & $-4.73$ & 8.1 \\
2 & \textit{NRXN1} & \textbf{Circuit} & \textbf{LTR} & \textbf{ERV3-16A3\_I} & $\mathbf{-2.53}$ & \textbf{1.9} \\
3 & \textit{EGFR} & Prolif. & LINE & L1MC1 & $-2.49$ & 7.9 \\
4 & \textit{CALB1} & Control & LTR & HERVH-int & $-2.46$ & 1.5 \\
5 & \textit{MBP} & Control & LINE & L1MD2 & $-2.42$ & 6.5 \\
6 & \textit{PLP1} & Control & LTR & HERVH-int & $-2.23$ & 8.2 \\
7 & \textit{PIK3CA} & Prolif. & LINE & L1PA15 & $-2.17$ & 8.9 \\
8 & \textit{MAP2} & Control & LINE & L1MC2 & $-2.07$ & 7.8 \\
9 & \textit{NLGN1} & \textbf{Circuit} & \textbf{LINE} & \textbf{L1PA7} & $\mathbf{-1.93}$ & \textbf{11.2} \\
10 & \textit{TERT} & Prolif. & LINE & L1MA3 & $-1.88$ & 8.2 \\
\bottomrule
\end{tabular}
\end{table*}

Beyond the top 10 overall (Table~\ref{tab:top_hits}), notable circuit-gene-specific hits include
\textit{SNAP25} L1PA6 ($\ris=-1.87$, 7.8\,kb),
\textit{THBS2} Tigger1 ($\ris=-1.69$, 3.7\,kb),
\textit{CREB1} L2a ($\ris=-1.36$, 6.3\,kb),
\textit{GRIA2} L1MB1 ($\ris=-1.31$, 9.5\,kb),
\textit{GAD2} L1MA8 ($\ris=-1.29$, 9.3\,kb),
\textit{NTRK2} L1ME1 ($\ris=-1.27$, 2.9\,kb),
\textit{GRIA4} L2c ($\ris=-1.25$, 8.8\,kb),
\textit{CREB1} L2b ($\ris=-1.22$, 1.9\,kb),
and \textit{SYT1} L1ME3E ($\ris=-1.07$, 7.9\,kb).
All are LINE or DNA transposon elements within the 10\,kb regulatory horizon.

\subsection*{Broader Impact Statement}

This work develops computational tools for characterizing noncoding regulatory elements in the context of glioma biology. The identified regulatory elements are computational predictions that require experimental validation before any clinical translation. We do not foresee negative societal impacts from this research, as it contributes to basic understanding of gene regulation in cancer. The methods and gene sets are derived from publicly available datasets (GENCODE, ENCODE, UCSC RepeatMasker) and publicly available pretrained models (Caduceus-Ph, HyenaDNA, Enformer). No patient data or human subjects were involved. The computational framework is generalizable and could accelerate noncoding variant interpretation in other diseases, potentially benefiting precision medicine efforts.


\end{document}